\documentclass{article}

\usepackage{arxiv,natbib}

\usepackage[utf8]{inputenc} 
\usepackage[T1]{fontenc}    
\usepackage{hyperref}       
\usepackage{url}            
\usepackage{booktabs}       
\usepackage{amsfonts}       
\usepackage{nicefrac}       
\usepackage{microtype}      
\usepackage{lipsum}
\usepackage{graphicx}
\usepackage{amsmath}
\usepackage{mathtools} 
\usepackage{latexsym}

\title{Automatic Data Expansion for Customer-care Spoken Language Understanding}

\author{
Shahab Jalalvand \\
Senior Research Scientist \\
Interactions Corp. \\
Murray Hill, NJ, USA \\
sjalalvand@interactions.com \\
\And
Andrej Ljolje \\ 
Principal Research Scientist \\
Interactions Corp. \\
Murray Hill, NJ, USA \\
aljolje@interactions.com \\
\And
Srinivas Bangalore \\
Director AI Research \\
Interactions Corp. \\
Murray Hill, NJ, USA \\
sbangalore@interactions.com \\
}

\begin{document}
\maketitle

\begin{abstract}

Spoken language understanding (SLU) systems are widely used in handling of customer-care calls.
A traditional SLU system consists of an acoustic model (AM) and a language model (LM) that are used to decode the utterance and a natural language understanding (NLU) model that predicts the intent.
While AM can be shared across different domains, LM and NLU models need to be trained specifically for every new task.
However, preparing enough data to train these models is prohibitively expensive.
In this paper, we introduce an efficient method to expand the limited in-domain data. 
The process starts with training a preliminary NLU model based on logistic regression on the in-domain data.
Since the features are based on $n=1,2$-grams, we can detect the most informative n-grams for each intent class. 
Using these n-grams,
we find the samples in the out-of-domain corpus that 1) contain the desired n-gram and/or 2) have similar intent label.
The ones which meet the first constraint are used to train a new LM model and the ones that meet both constraints are used to train a new NLU model.
Our results on two divergent experimental setups show that the proposed approach reduces by 30\% the absolute classification error rate (CER) comparing to the preliminary models and it significantly outperforms the traditional data expansion algorithms such as the ones based on semi-supervised learning, TF-IDF and embedding vectors.

\end{abstract}

\section{Introduction}

The usage of spoken language understanding (SLU) in costumer-care applications is increasing everyday.
Traditional SLU system 
consists of a pipeline of automatic speech recognition (ASR) and natural language understanding (NLU) \cite{huang2001spoken}.
ASR has two main components: an acoustic model (AM) and a language model (LM) \cite{rabiner1993fundamentals}.  
AM is trained on the acoustic information and LM is trained on the text corpora.
While AM can be shared across different domains, LM and NLU models need to be trained specifically for every new task.
Assuming that AM is generic enough to be used across different tasks, in this paper, we focus on the LM and NLU components and we address two main issues:

\begin{enumerate}
    \item \textbf{LM is usually trained independently from NLU model.} Therefore it does not receive any feedback from the NLU model about the important words (or n-grams) in the task.
    \item \textbf{The availability of in-domain data is usually very limited.} Therefore training accurate LM and NLU models is not possible on day zero.
\end{enumerate}

Addressing the first issue, \cite{yaman2008integrative} proposes an integrative and discriminative technique to update the parameters of the LM and NLU models.
In this technique, the n-best hypotheses that are generated by the ASR decoder are rescored and reranked using the AM, LM and NLU scores. 
Then the best hypothesis (i.e. the one whose intent is truly predicted with the highest score) and the most competitive hypothesis (i.e. the one whose intent is predicted wrongly with high score) are detected and used to update the n-gram probabilities in LM and NLU.

To solve the second issue, there are many data expansion techniques in the literature. 
These techniques can be categorized into two main categories: generative and selective techniques. 
In generative techniques a generative model like recurrent neural network language model (RNNLM) \cite{mikolov2011rnnlm} is trained on the in-domain data and then the model is used to generate similar samples \cite{bowman2015generating, tam2015rnn}.
The selective approaches are mostly based on searching through a big out-of-domain (OOD) corpus and finding most similar samples according to an appropriate similarity measures.
Some of these methods are based on cross entropy \cite{moore2010intelligent}, term frequency inverse document frequency (TF-IDF) \cite{salton1987term} and word embeddings \cite{mikolov2013distributed}.

This paper addresses the two mentioned issues and presents an automatic data expansion algorithm that: 
\textbf{
\begin{itemize}
    \item Takes advantage of a preliminary NLU model to detect the most informative words (or n-grams) and
    \item Uses those n-grams to select relevant data from an out-of-domain corpus.
\end{itemize}
}

The process starts with training a preliminary NLU model based on logistic regression on the in-domain data.
Since the features are based on $n=1,2$-grams, we can detect the most informative n-grams for each intent class. 
Therefore, for each intent
we prepare a list of representative n-grams by excavating the trained feature weight matrix in the NLU model.
Table \ref{tab:intentexampleA} shows some examples of in-domain sentences and the obtained list of n-grams.
\begin{table*}[h]
    \centering
    \begin{tabular}{l|c||l}
        Audio Utterance & Intent & n-gram list \\ \hline\hline
        i'm calling about charges & BILLING & (charges, paying) \\
        uh password help & ACCOUNT & (pass code, password, my account) \\
        making appointment & APPOINTMENT & (reservation, reservations) \\
        complain about delivery	& COMPLAINT & (complain, bad experience)\\\hline
    \end{tabular}
    \caption{in-domain data examples for task \textit{A}.}
    \label{tab:intentexampleA}
\end{table*}
For every n-gram, we find samples in the OOD corpus that 1) contain the desired n-gram and/or 2) have similar intent label.
The OOD samples that meet the first constraint are used to train the new LM model and the ones that meet both constraints are used to train the new NLU model.
Our results on two divergent 
tasks
show that the proposed approach reduces the classification error rate (CER) by 30\% comparing to the preliminary models and it significantly outperforms the traditional data expansion algorithms such as the ones based on semi-supervised learning \cite{asli2013semi}, TF-IDF \cite{salton1987term} and embedding approaches \cite{mikolov2013distributed}.

The rest of this paper is organized as follows.
Section 2 describes the related work. 
Section 3 briefly describes the different SLU components;
Section 4 describes the proposed automatic data expansion approaches; 
Section 5 contains the experimental setup and results
and finally, Section 6 concludes this paper.

\section{Related Work}

Scientific literature related to the work described in this article spans over two main lines of investigation: a) integrative training of LM and NLU models and b) automatic data expansion.

\textbf{Integrative training of LM and NLU.}
\cite{yaman2008integrative} proposes an integrative and discriminative technique to update the parameters of LM and NLU models.
In this technique, the n-best list hypotheses that are generated by an ASR decoder are first rescored and reranked using the AM, LM and NLU scores. 
Then the best hypothesis (i.e. the one whose intent is truly predicted with the highest score) and the most competitive hypothesis (i.e. the one whose intent is predicted wrongly with high score) are detected.
Finally, these discriminant hypotheses are used to update the
probability of the n-grams 
in both the LM and NLU models.
The performance of this approach is highly dependent on the performance of the preliminary LM which is used to generate the n-best list.
Moreover, in order to optimize the weights of the AM, LM and NLU scores, a lot of in-domain data is required. 
Our proposed approach also exploits the NLU information to update the LM and NLU even though very limited in-domain data is available. 

Other researches try to remove the need of ASR in the SLU system and build an end-to-end deep neural network that directly predicts the intent from acoustic information \cite{price2018spoken}.
Although they have obtained competitive results, these methods have some deficiencies that prevent them to be used in the industry. 
For example whenever a new intent is introduced, the whole network need to be retrained. 

\textbf{Data expansion.} 
An intelligent method to expand the LM training data is proposed in \cite{moore2010intelligent}. 
In this method, two language models are trained: one on in-domain and the other on out-of-domain data.
Then all the out-of-domain sentences are measured by a criterion that is based on cross entropy difference between the in-domain and out-of-domain LMs. 
The larger the difference, the higher the chance to be selected.
This approach however needs a reasonable size of in-domain data to train the in-domain LM, while as mentioned before, our available data is as small as several hundred utterances.

To expand data for NLU models, one solution is through semi-supervised learning \cite{nigam2000text}.
In this approach, a preliminary NLU model is trained on the in-domain data and then it is used to label a large number of unlabeled data.
The new NLU model is trained on the predicted labels.
However, as we will see in our experiments, the preliminary model can be such poor that it fails to assign proper intent labels to the unlabelled utterances.

\section{Spoken Language Understanding}
\begin{figure}[h]
\centering
\includegraphics[trim=0cm 0cm 0cm 0cm, clip=true, width=0.75\linewidth]{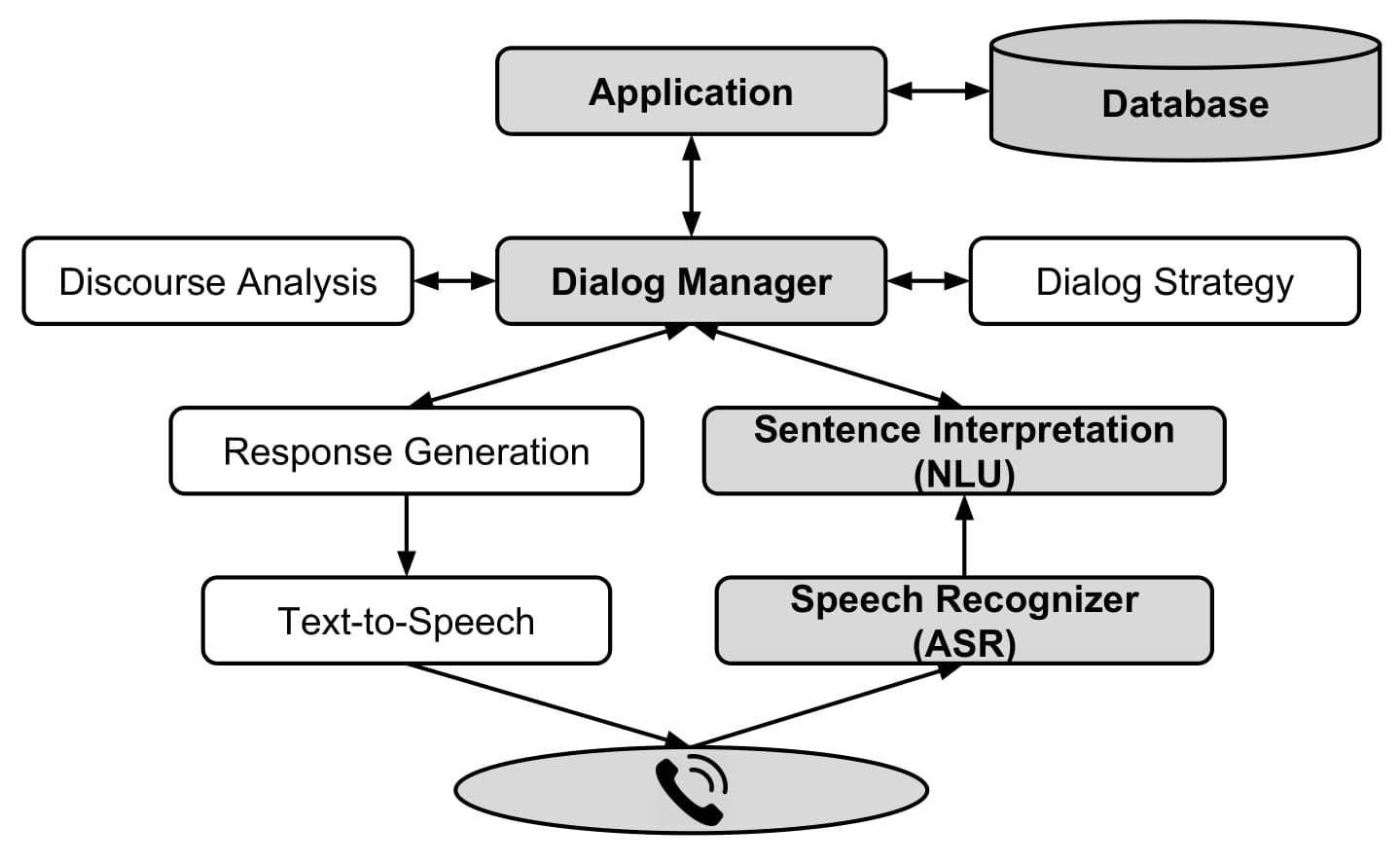}
\caption{Basic SLU system architecture for costumer-care calls.}
\label{fig:slu}
\end{figure}

Figure \ref{fig:slu} shows a block diagram of a costumer-care SLU system, highlighting the modules that we focus in this paper \ref{fig:slu}.
A transaction starts with a prompt like: 

\noindent
$How\ may\ I\ help\ you?$. 

\noindent
The human responds with usually short utterances: 

\noindent
$I\ want\ to\ make\ an\ appointment$.

\noindent
ASR generates a hypothesis for this utterance 
and NLU predicts the intent label for this hypothesis as:

\noindent
$APPOINTMENT$.

Based on the predicted intent label and its confidence score, \textit{Dialog Manager} provides the proper response for this request and the dialog continues.

\subsection{Automatic speech recognition (ASR)}
ASR aims to find the most probable sequence of words, $\hat{W}$, given a sequence of acoustic observations, $X$:
\begin{flalign}\nonumber
\label{eq:asr}
& \hat{W} = \underset{W} {argmax} \{P^{1/L}[X|W] \times P[W]\} 
\end{flalign}

\noindent
where $X=x_{1},x_{2},...,x_{T}$ indicates the sequence of acoustic observations;
$P[X|W]$ computes the acoustic model likelihood and
$P[W]$ computes the language model likelihood. 
Since the AM and LM scores have different dynamic range, an LM scale factor $L$ is used to balance the scores.

Since in this paper, we focus on LM and NLU models, we assume that AM is unchanged from one experiment to the other.

Our language model is a traditional back-off n-gram LM \cite{chen1999empirical}.
Thanks to its simple structure, fast training and easy implementation, n-gram LM has been widely used in the industry.
In a back-off n-gram LM, the conditional probability of a word $w_i$ given the $n-1$ previous words $w_{i-n+1}^{i-1}$ is computed by:
\begin{flalign}\nonumber
& P_{bo}(w_{i} | w_{i-n+1}^{i-1}) =  \\
& \left\{\begin{matrix} 
  d_{w_{i-(n-1)}^{i}} \times \frac{C(w_{i-(n-1)}^{i})}{C(w_{i-(n-1)}^{i-1})},\ \ \ if\ C(w_{i-(n-1)}^{i}) >  k
  \\ 
  \alpha_{w_{i-(n-1)}^{i-1}} \times P_{bo} (w_i| w_{i-(n-2)}^{i-1}),\ \ \ otherwise
\end{matrix}\right.
\end{flalign}

In this formula, $p_{bo}(w_{i} | w_{i-n+1}^{i-1})$ is the back-off probability of observing $w_i$; 
$C(w)$ is the frequency of $w$ in training set; 
$k$ is a threshold for the least acceptable number of appearances and 
$d$ is the Good Turing discounting estimation.
Other extensions such as modified Kneser-Ney also called modified shift-beta smoothing have shown very good performance \cite{chen1999empirical}.

Our proposed data expansion algorithm modifies the probability of the specific n-grams that contribute the most in intent prediction.
This modification is done by adding more hypotheses that contain those specific n-grams.

\subsection{Natural language understanding (NLU)}
An NLU model classifies the hypothesis $\hat W$ into one of $L$ intent classes $\hat Y \in \{Y_1 ... Y_L\}$.
In SLU, $\hat W$ is generated by the ASR decoder, Eq. \ref{eq:asr}.
To achieve a more robust intent prediction, instead of a single hypothesis $\hat W$, the n-best lists are used.

We use binary n-grams with $n=1,2$ to form the feature vector for a recognized hypothesis.
The dimension of the feature vector is equal to the total number of uni- and bi-grams in the training data.
$k$-th element of the vector is $1$, if its corresponding n-gram is seen in the hypothesis.

For the classifier, we use logistic regression with hinge loss, averaged stochastic gradient descent (SGD) and $L2$ feature normalization.
The objective function for training is:
\begin{equation}
    \underset{M,b}{min} \frac{1}{N} \sum^{N}_{i=1} \xi_i . Loss(y_i, M^Tx_i + b) + \frac{\alpha}{2} \left \| M \right \|_2^2
\end{equation}

\noindent
where $M$ is the estimated weight matrix; 
$b$ is the bias; 
$N$ is the total number of training samples;
$y_i$ is the label for the $i$-th training sample and 
$x_i$ is the feature vector for the $i$-th training sample.

Matrix $M$ is $[I\times L]$ dimensional, in which $I$ is the total number of uni- and bi-grams and $L$ is the total number of intent classes.
\textbf{Our proposed data expansion method makes use of this matrix as a source of information to identify the most informative n-grams for each intent.}

We are aware of other classifiers based on deep neural networks like the ones in \cite{collobert2008unified, zhang2015character, goldberg2016primer}, however we use simple logistic regression because we are interested in evaluating different data expansion techniques. 
Analyzing the classifier performance is beyond the context of this paper.

\section{Data Expansion Approaches}
In this section, we describe four techniques for data expansion.

\subsection{Semi-supervised learning approach}
In this approach, an NLU model is trained on the in-domain data and then it is used to assign an intent label to the utterances  in the OOD corpus.
The samples with high confidence score are used to train the second NLU model.
This process is done for several iterations until convergence to an optimal model \cite{nigam2000text}.


\subsection{TF-IDF approach}
Another strategy to select relevant data is using 
similarity measures.
One of the most popular methods to compute sentence similarity is through term frequency-inverse document frequency (TF-IDF).
Similarity of the two sentences can be computed as cosine similarity between the TF-IDF vectors of the two sentences.
\begin{flalign}
\label{eq:tfidf}
& cos\_sim(s_1,s_2) = &&\\ \nonumber
& \ \ \ \ \ tfidfVect(s_1).tfidfVect(s_2) \\\nonumber \\ \nonumber
& tfidfVect(s) = \\ \nonumber
& \ \ \ \ \ [tfidf(w_1,s), tfidf(w_2,S),...] \\\nonumber \\ \nonumber
& tfidf(w,s) = tf(w,s) \times idf(w) \\\nonumber \\ \nonumber
& idf(w) = 1+log( \frac{N}{ |\{s:w\in s\}| } )
\end{flalign}

\noindent
where, $N$ is the total number of sentences and $tf(w,s)$ is the frequency of $w$ in sentence $s$.

In the experiments, 
we collect all the in-domain and out-of-domain utterances into a large document and we compute a TF-IDF vector for each sentence. 
Then the cosine similarity is computed between every in-domain and all out-of-domain samples.
For every in-domain sample, the most similar OOD samples are selected to train the new LM and NLU models.


\subsection{Embedding approach}
Continuous space embeddings such as word2vec \cite{mikolov2013distributed} project terms to a continuous and dense feature space.
When the words and sentences are represented by a continuous feature vectors, their similarity can be measured using simple metrics such as euclidean distance.

We apply embedding methods to expand our limited in-domain data.
To do this, a word2vec model is trained on the whole available sentences including in-domain and out-of-domain corpora.
This model represents each word with a continuous feature vector.
In order to extend the word vectors to sentence vectors, we use average pooling. 
Although average pooling is not the most efficient approach, it is a common way to convert word vectors into sentence vectors.
For every in-domain sentence, we find the most similar ones in the OOD corpus, using euclidean distance.
Among the selected sentences, the ones whose intent label matches the desired in-domain sample will be selected for the NLU training and the rest for the LM training.

\subsection{N-gram based data selection}
\label{nlu-driven}

\begin{figure*}[t]
\centering
\includegraphics[trim=0cm 0cm 0cm 0cm, clip=true,width=\linewidth]{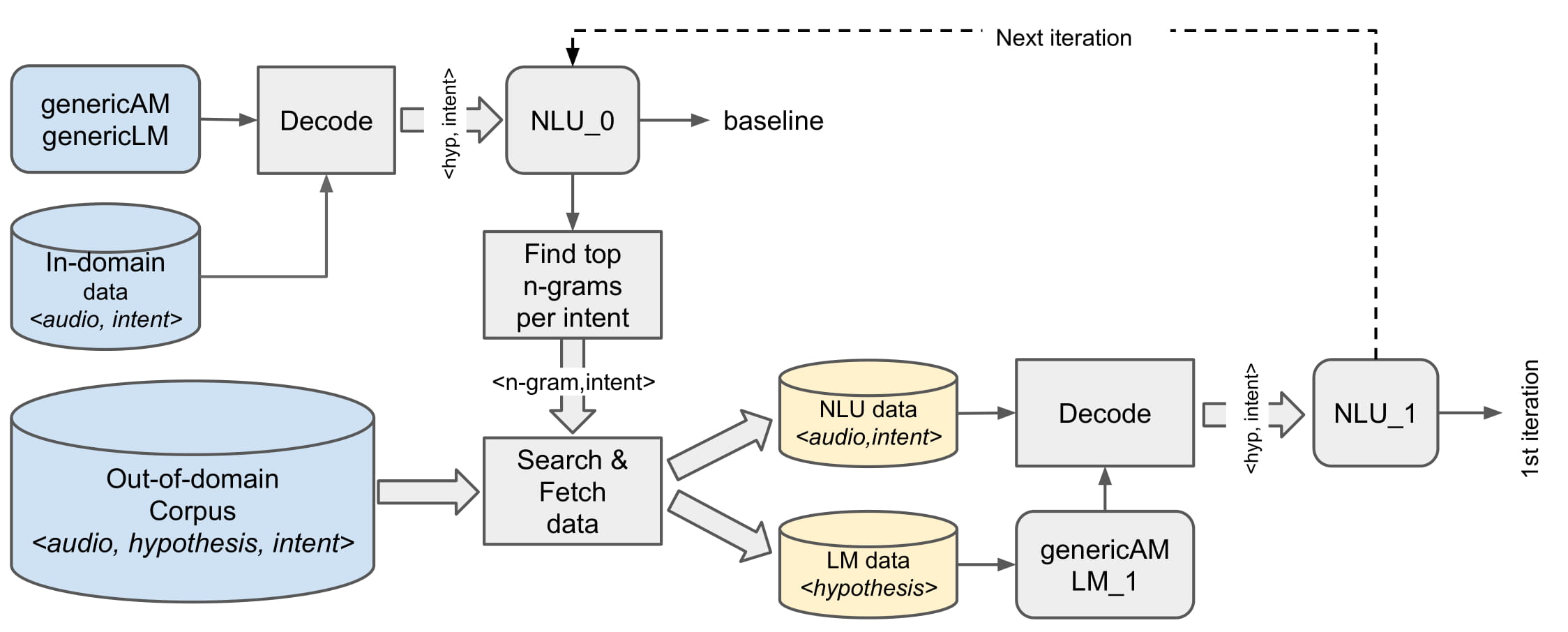}
\caption{NLU-driven data expansion architecture.}
\label{fig:nlu-driven}
\end{figure*}

Figure \ref{fig:nlu-driven} shows our proposed algorithm for data expansion using n-gram features. 
The available resources are: 
\begin{itemize}
    \item \textit{out-of-domain (OOD) corpus} that is a large corpus containing ~58 million utterances from 29 costumer care applications. Each utterance comes with an audio, a recognized hypothesis and an intent label.
    \item \textit{genericAM} that is an acoustic model trained on hundreds of telephone calls from the 29 applications.
    \item \textit{genericLM} that is an interpolated language model between 29 LMs, each trained for one application.
    \item \textit{in-domain data} that contains several utterance examples per intent from the new application.
\end{itemize}

We first use \textit{genericAM} and \textit{genericLM} to decode the in-domain utterances.
Then, \textit{NLU\_0} is trained on the $<hypotheses,intent>$ pairs 
using logistic regression with $n=1,2$-grams as features.
The trained weight matrix in \textit{NLU\_0} is excavated to detect and create a list of the most informative n-grams for each intent (see Table \ref{tab:intentexampleA} for examples).
These n-grams are defined as the ones with highest positive weights or the ones with lowest negative weights or a combination of both.
For each n-gram, we find utterances in \textit{OOD corpus} that meet two constraints: 1) contain the desired n-gram and 2) has similar intent label.
The ones that meet the first constraint are used to train \textit{LM\_1}.
The ones that meet both constraints are used to train \textit{NLU\_1}.
Note that the sentences in NLU training set are a subset of LM training.

\section{Experimental Setup}
We evaluate the performance of our data expansion algorithm with three alternative approaches:

\begin{itemize}
    \item semi-supervised approach;
    \item TF-IDF based data selection;
    \item embedding approach.
\end{itemize}

We conduct the experiments in two divergent circumstances:
\begin{enumerate}
    \item \textbf{Task\_A: high intent coverage}. When most of the intent labels in the new application \textit{task\_A} already exist in the OOD corpus.
    \item \textbf{Task\_B: low intent coverage}. When most of the new intent labels do not exist in the OOD corpus.
\end{enumerate}

Since \textit{genericAM}, \textit{genericLM} and \textit{OOD corpus} are shared across the two tasks, we describe them beforehand.

\textbf{\textit{genericAM}}.
Our generic acoustic model is a hybrid DNN-HMM model \cite{hinton2012deep} trained with the cross-entropy criterion followed by the state-level Minimum Bayes Risk (sMBR) objective.
The training set consists of about 380 hours of transcribed utterances.

\textbf{\textit{genericLM}}.
The generic language model that we use in the first pass of decoding is a 3-gram interpolated language model between 29 individual LMs.
Each individual LM corresponds to one application and is trained using 3-gram Katz's back-off \cite{katz1987estimation}. 
The largest and smallest LMs are trained on 588 million and 8 million words, respectively. 


\textbf{\textit{Out-of-domain (OOD) corpus}}.
This corpus consists of the utterances from 29 existing applications (excluding application \textit{A} and \textit{B}).
Table \ref{tab:oodcorpus} shows the statistics of this corpus.

\begin{table}[h]
    \centering
    \begin{tabular}{l|c}
    Description & Stats \\ \hline \hline 
    number of applications & 29 \\
    number of utterances & 58 M \\
    number of words & 202 M \\
    size of vocabulary & 23 K \\
    number of unique intents & 3392 \\ \hline
    \end{tabular}
    \caption{Statistics of the out-of-domain corpus.}
    \label{tab:oodcorpus}
\end{table}

\subsection{Task\_A}
In this task, the new intent labels for the new application \textit{A} are well covered in \textit{OOD corpus}.
About 93\% of the new intents are already used in the previous applications.

\textbf{In-domain data A.}
Application \textit{A} has 78 different intents and for each intent only 10 audio samples are provided. 
Therefore, in total there are 780 $<audio, intent>$ samples. 
In Table \ref{tab:intentexampleA}, we showed some examples of these utterances.

\textbf{Test set A.}
The test set of this application consists of 4891 customer care calls that are transcribed and labeled by human transcriber.
The whole set is ~7 hours, including ~28 K words and there are 78 unique intent labels.
It's worth remembering that 93\% of these intent labels are covered in \textit{OOD corups}.

\begin{table*}[h]
    \centering
    \begin{tabular}{l|l|c|c}
 LM (\# words) & NLU (\# utterances) & WER[\%]   & CER[\%] \\ \hline \hline
 genericLM & inNLU (780) & 17.1  & 61.43 \\ \hline
 genericLM & in+semi-supNLU1 (60 k) & 17.1  & 57.31 \\ 
 genericLM & in+semi-supNLU2 (60 k) & 17.1  & 54.63 \\ \hline
 genericLM & in+tfidfNLU1 (60k) & 17.1  & 52.52 \\
 tfidfLM (2 M) & in+tfidfNLU (60k) & 26.1  & 51.65 \\ \hline
 genericLM & in+embedNLU (60k) & 17.1 & 53.67 \\
 embedLM (2 M) & in+embedNLU (60k) & 19.9 & 51.31 \\ \hline
 genericLM & in+ngselNLU1 (60 k)  & 17.1  & 49.98 \\
 ngselLM1 (2 M) & in+ngselNLU1 (60 k)  & 18.5  & 49.35 \\
 ngselLM2 (2 M) & in+ngselNLU2 (60 k)  & 18.7  & 48.43 \\ \hline
    \end{tabular}
    \caption{The performance of different models with different data selection approaches in task \textit{A}.}
    \label{tab:resultsA}
\end{table*}

\textbf{Results A.} Table \ref{tab:resultsA} shows the word error rate (WER[\%]) and classification error rate (CER[\%]) results by different data expansion approaches.


In the first row, we  use \textit{genericLM} for recognition.
For intent prediction, we use \textit{inNLU} which is trained on the in-domain data only. 
The WER result is 17.1\% and the CER is 61.43\%.
This is the setup that we consider as baseline.

The next two rows report the performance of the semi-supervised approach in two iterations. 
In the first iteration, \textit{inNLU} is used to label all the utterances in \textit{OOD corpus}.
The ones with high confidence score (together with the in-domain data) are used to train \textit{in+semi-supNLU1}.
We tune the confidence threshold, so that we can retrieve a reasonable number of 60 K samples.
The reason for selecting 60 K is simply for the sake of consistency in comparison between different data selection approaches.
In the second iteration, we use \textit{in+semi-supNLU1} to label the OOD samples.
We observe that semi-supervised approach improves CER to 57.31\% and 54.63\%, respectively, in the first and second iterations.

The third set of rows reports the performance of TF-IDF based approach.
For each in-domain sample, we keep the 10 K most similar samples in the OOD corpus according to Eq. \ref{eq:tfidf}.
This number provides the desired 60 K data to train
\textit{in+tfidfNLU}.
This model yields 52.52\% CER. 
This approach can be used to collect more data (i.e. by removing the constraint of intent label match) for training a new LM. 
This data along with the ones for NLU training are used to build \textit{tfidfLM}.
This new LM yields worse WER results 26.1\%, though it improves the CER to 51.65\%.
Such a strange behaviour, i.e. WER increase and CER decrease, may happen when LM is trained together with NLU
\cite{wang2003word, yaman2008integrative}.

The fourth set of results are obtained by embedding approaches.
Using embedding vectors with some thresholds on the euclidean distance between in-domain and out-of-domain sentence vectors, we collect 60 K utterances to train \textit{in+embedNLU} model. 
This model yields 53.67\% CER which is 1.05\% worse than TF-IDF approach when \textit{genericLM} is used to decode the utterances.
Then, we train \textit{embedLM} on the data selected by embedding approach. 
This LM results in a much better WER performance (19.9\%) comparing to its TF-IDF counterpart (26.1\%) and it also outperforms TF-IDF method in terms of CER. 

In the last set of rows, we evaluate our proposed data expansion algorithm by using n-gram features  
from \textit{inNLU}.
How many n-grams per intent and how many samples per n-gram?
These are the parameters that are again tuned to retrieve about 60 K instances to train \textit{ngselNLU1}.
Using this  model along with the \textit{genericLM}, we obtain 49.98\% CER.
This result shows an absolute improvement of 2.54\% CER in comparison to the TF-IDF method (52.52\%) and 3.69\% in comparison to the embedding approach. 
Again, this method can be used to select data for training a customized LM \textit{ngselLM1}.
This LM although increases the WER to 18.5\%, it slightly reduces the CER to 49.35\%.
In the second iteration, we use the same approach, this time by using n-gram features from \textit{in+ngselNLU1}.
The selected data are used to train \textit{ngselLM2} and \textit{ngselNLU2} yielding
48.43\% CER.

\begin{figure}[h]
\centering
\includegraphics[trim=0cm 0cm 0cm 2cm, clip=true,width=0.75\linewidth]{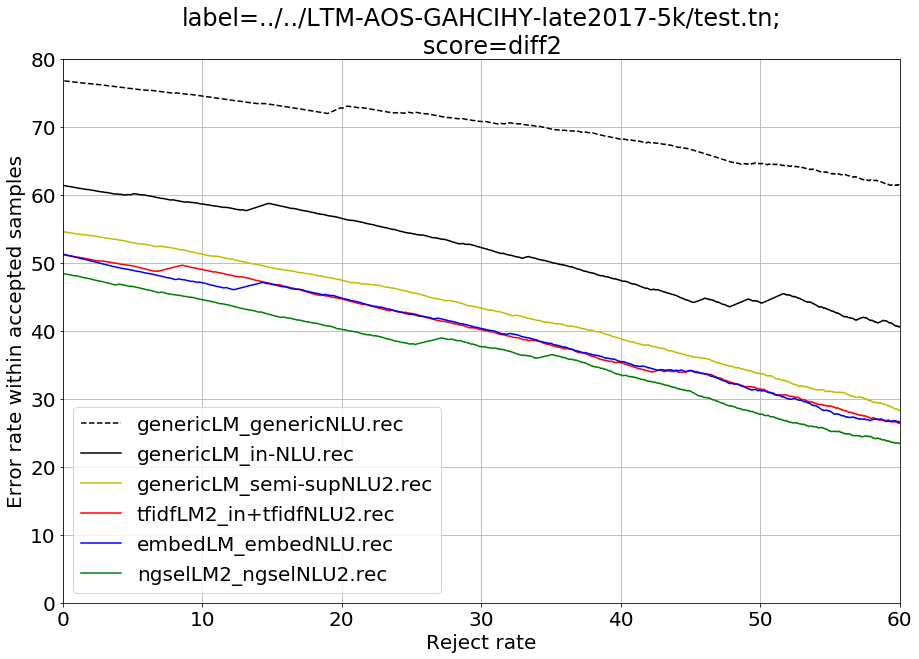}
\caption{Error rate within accepted samples as a function of rejection rate (Lift curve).}
\label{fig:certaskA}
\end{figure}

Figure \ref{fig:certaskA} shows the Lift curves \cite{tuffery2011data} obtained by different selection strategies.
Lift curve shows the error rate within accepted samples as a function of rejection rate that is applied on the confidence score of the classifier. 
As we can see in this figure, the proposed n-gram based data selection approach consistently outperforms the other approaches in all rejection rates.

To investigate why n-gram based data expansion outperformed TF-IDF and embedding approaches, we look at the diversity of the data selected by each method.
The vocabulary size of the data selected by TF-IDF, embedding and n-gram based data selection is 2169, 2572 and 3030 words, respectively.
\textbf{That is, n-gram based data expansion is able to find more variety of data, whereas, TF-IDF and embedding, due to the way they work,  try to find the most similar sentences to the in-domain data, ignoring the fact that some words (or n-grams) are more important to predict the intent of a hypothesis.}

\subsection{Task\_B}
In this task, the new application \textit{B} has a lot of unseen intent labels.
Only 25\% of the intents in application \textit{B} is previously seen in the OOD corpus.
Our goal is to verify the performance of our data selection algorithm in a challenging scenario, where there are a lot of new intent labels.

\begin{table*}[h]
    \centering
    \begin{tabular}{l|c||l|c}
        \multicolumn{2}{c||}{In-domain Examples} & \multicolumn{2}{c}{Out-of-domain Examples} \\
        Utterance & Intent & Utterance & Intent \\ \hline\hline
        credit card bill & BILLING & account charges & BILLING \\
        i need my account unlocked  & ACCOUNT & trouble logging in & ACCOUNT \\
        problem with the website & ONLINE\_WEB\_HELP & online support & WEB\_HELP \\
        i need update for version two & UPDATE\_SOFTWARE & edit my name & UPDATE\_PROFILE \\
        remove a form & FORMS & -- & -- \\ \hline
    \end{tabular}
    \caption{Left: in-domain data examples for task \textit{B}. Right: the matched intent examples from out-of-domain corpus. }
    \label{tab:intentexampleB}
\end{table*}

\textbf{In-domain data B.}
The provided in-domain data for this task has 91 different intents.
Again we consider having only 10 audio samples per intent. 
Therefore, in total there are 910 $<audio, intent>$ samples in the in-domain data.
Table \ref{tab:intentexampleB} shows some examples.

\textbf{Test set B.}
The test set contains 3657 costumer-care utterances which is about 6.5 hours.
For this set, there is no human transcription available, so that we are not able to evaluate WER results.
However, the true intent labels are available for all test utterances.
There are 100 different labels and only 25\% of the labels are seen in \textit{OOD corpus}.

\textbf{Results B.}
Our first observation about the unseen intent labels is that, although there are many unseen labels, for many of them we can find similar labels. 
Therefore, as a naive solution for this problem, we make use of string similarity to find intent labels for the unseen ones.
To do this, we use python's \textit{difflib} library and its \textit{get\_close\_matches} function to find the closest matches for the unseen labels.
As we see in Table \ref{tab:intentexampleB} sometimes we can find exact matches in \textit{OOD corpus}, sometimes we find close matches and sometimes there is no match.

\begin{table}[h]
    \centering
    \begin{tabular}{l|l|c}
 LM (\# words) & NLU (\# utterances) & CER[\%] \\ \hline \hline
 genericLM & inNLU & 71.28 \\ \hline
 ngselLM (3 M) & in+ngselNLU (48 k) & 44.21 \\ \hline
    \end{tabular}
    \caption{The performance of the proposed n-gram-based data expansion in task \textit{B}.}
    \label{tab:resultsB}
\end{table}

From Table \ref{tab:resultsB}, 
again we observe a huge improvement from 71.28\% to 44.21\% in CER. 
This result already verifies the effectiveness of data expansion using n-gram features.
It worth analyzing the improvement gap as a function of in-domain data size.

\subsection{Improvement wrt. in-domain data size}
In the further analyses, we are interested to know, to what extent our data expansion approach improves the CER results when there is more in-domain data.
In the previous experiments, we assumed that only 10 samples per intent were available.
In the next experiments, we enlarge this number to 50, 100 and 150 and
we conduct experiments on task \textit{B}.

\begin{figure}[h]
\centering
\includegraphics[trim=0cm 0cm 0cm 2cm, clip=true,width=0.75\linewidth]{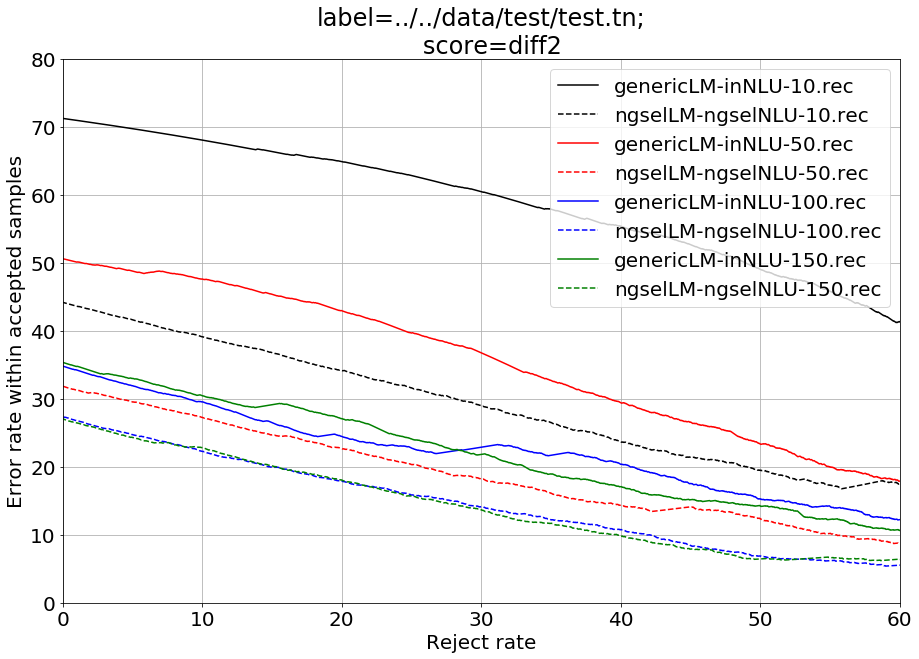}
\caption{Improvement gap as a function of in-domain data size (i.e. 10, 50, 100 and 150 samples per intent).}
\label{fig:indomainsize}
\end{figure}

Figure \ref{fig:indomainsize} shows the Lift curve of models trained only on in-domain data (solid lines) and the ones trained on selected data using n-gram based data expansion (dotted lines).
As it can be seen in Figure \ref{fig:indomainsize}, by enlarging the in-domain data to 50 and 100 samples per intent, the baseline models get improved consistently. 
From 100 to 150 samples per intent, we see a slight increase in CER.
Regardless to the size of in-domain data, n-gram based data expansion significantly improves the CER results.

\section{Conclusion}
We introduced a novel technique to expand the limited in-domain data for a new spoken language understanding (SLU) task.
This technique takes advantage of the key information that a preliminary NLU model (trained only on in-domain data) provides and it uses this information to find more relevant data from an out-of-domain corpus.
Using this method we were able to significantly enlarge the training data for both language model and NLU model.
The new models showed significantly better performance in terms of classification error rate, in two divergent experimental setup.

Future work is dedicated to a) designing a more efficient algorithm to match the unseen intent labels and b) preparing the out-of-domain corpus using clustering approaches.


\bibliographystyle{plain}
\bibliography{references}

\end{document}